\documentclass[sigconf]{acmart}

\usepackage{booktabs} 
\usepackage{comment}
\usepackage{soul}
\usepackage{caption}
\usepackage{float}
\usepackage{amsfonts}
\usepackage[normalem]{ulem}
\usepackage{multirow}
\usepackage{hyperref}
\usepackage{amsfonts}
\usepackage{amsmath,bm}
\usepackage{amsthm}
\usepackage{times}

\usepackage{algorithm}

\usepackage[noend]{algpseudocode}
\usepackage{color}
\usepackage{array,booktabs}

\setcopyright{none}

\acmConference[SIGIR'19]{The 42nd International ACM SIGIR Conference on Research \& Development in Information Retrieval}{July 21-25, 2019}{Paris, France}
\acmBooktitle{SIGIR'19: The 42nd International ACM SIGIR Conference on Research \& Development in Information Retrieval, July 21-25, 2019, Paris, France}
\copyrightyear{2019}
\acmYear{2019}

\acmPrice{}
\newcommand{\bu}{{\mathbf{u}}}
\newcommand{\bv}{{\mathbf{v}}}
\newcommand{\bw}{{\bm{w}}}
\newcommand{\id}{{\mathbf{i}}}

\newcommand{\uid}{\bu_{[\id]}}

\newcommand{\btheta}{{\bm{\theta}}}
\newcommand{\phiii}{\phi^\init_{[\id]}}
\newcommand{\phiprime}{\phi'_{[\id]}}

\newcommand{\newid}{{\mathbf{i^*}}}

\newcommand{\unewid}{\bu_{[\newid]}}
\newcommand{\phiiinew}{\phi^\init_{[\newid]}}

\newcommand{\init}{{\textrm{init}}}

\newcommand{\meta}{\textrm{meta}}

\begin{document}
\title{Warm Up Cold-start Advertisements: Improving CTR Predictions via Learning to Learn ID Embeddings}

\author{Feiyang Pan}
\affiliation{
  \institution{Institute of Computing Technology, Chinese Academy of Sciences}
}
\authornote{Corresponding author (panfeiyang@ict.ac.cn).}
\authornote{At the Key Lab of Intelligent Information Processing of Chinese Academy of Sciences. Also at University of Chinese Academy of Sciences, China. }

\author{Shuokai Li}
\affiliation{%
\institution{Institute of Computing Technology, Chinese Academy of Sciences}
}
\authornotemark[2]

\author{Xiang Ao}
\affiliation{%
\institution{Institute of Computing Technology, Chinese Academy of Sciences}
}
\authornotemark[2]
\author{Pingzhong Tang}
\affiliation{%
 \institution{IIIS, Tsinghua University}
}

\author{Qing He}
\affiliation{%
 \institution{Institute of Computing Technology, Chinese Academy of Sciences}
}
\authornotemark[2]
\begin{abstract}
Click-through rate (CTR) prediction has been one of the most central problems in computational advertising. Lately, embedding techniques that produce low-dimensional representations of ad IDs drastically improve CTR prediction accuracies. However, such learning techniques are data demanding and work poorly on new ads with little logging data, which is known as the cold-start problem.

In this paper, we aim to improve CTR predictions during both the \emph{cold-start} phase and the \emph{warm-up} phase when a new ad is added to the candidate pool. We propose \emph{Meta-Embedding}, a meta-learning-based approach that learns to generate desirable initial embeddings for new ad IDs. The proposed method trains an embedding generator for new ad IDs by making use of previously learned ads through gradient-based meta-learning. In other words, our method learns how to learn better embeddings. When a new ad comes, the trained generator initializes the embedding of its ID by feeding its contents and attributes. Next, the generated embedding can speed up the model fitting during the warm-up phase when a few labeled examples are available, compared to the existing initialization methods.

Experimental results on three real-world datasets showed that Meta-Embedding can significantly improve both the cold-start and warm-up performances for six existing CTR prediction models, ranging from lightweight models such as Factorization Machines to complicated deep models such as PNN and DeepFM. All of the above apply to conversion rate (CVR) predictions as well.

\end{abstract}
%
%

\maketitle

\section{Introduction}

The essence of online advertising is for the publisher to allocate ad slots in a way that maximizes social welfare. In other words, an ideal ad auction mechanism allocates, for each user, the best slot to the best ad, the second slot to the second best ad and so on. In order to determine which ad is the most valuable to this user, a key parameter is the so-called {\em click-through rate} also known as the CTR. A typical ad auction mechanism then allocates according to a descending order of the product of bid times its CTR. Since the bids are inputs from the advertisers, the main effort of the publisher is to estimate an accurate CTR for each ad-user pair in order to ensure optimal allocation. The same argument applies for conversion rate (CVR) if the publisher cares about how much these clicks successfully turn into business transactions. As a result, academia and major internet companies invest a considerable amount of research and engineering efforts on training a good CTR estimator.

A leading direction for predicting CTR has been the one that makes use of recent progress on deep learning~\cite{cheng2016wide,qu2016product,guo2017deepfm,pan2018field,zhou2018deep}. These deep models can be typically decomposed into two parts: Embeddings and Multi-layer Perceptrons (MLP)~\cite{zhou2018deep}. First, an embedding layer transforms each field of raw input into a fixed-length real-valued vector. Typical usage of the embedding is to transform an ad identifier (ad ID) into dense vectors, which can be viewed as a latent representation of the specific ad. It has been widely known in the industry that a well-learned embedding for an ad ID can largely improve the prediction accuracy, compared to methods with no ID input~\cite{cheng2016wide,juan2017field,qu2016product,guo2017deepfm,zhou2018deep}. Next, embeddings are fed into sophisticated models, which can be seen as different types of MLP. These models include Factorization Machine (FM)~\cite{rendle2010factorization,rendle2011fast}, the extended family of FFM \cite{juan2017field} and FwFM \cite{pan2018field} that use inner products of embeddings to learn feature interactions; deeper models such as Wide\&Deep~\cite{cheng2016wide}, PNN~\cite{qu2016product} and DeepFM~\cite{guo2017deepfm} that learn higher-order relations among features. These methods have achieved state-of-the-art performance across a wide range of CTR prediction tasks.

Despite the remarkable success of these methods, it is extremely data demanding to learn the embedding vectors.
For each ad, a large number of labeled examples are needed to train a reasonable ID embedding. When a new ad is added to the candidate pool, it is unlikely to have a good embedding vector for its ID with the mentioned methods. Moreover, for ``small'' ads with a relatively small number of training samples, it is hard to train their embeddings as good as for the ``large'' ads. These difficulties can all be regarded as the \emph{cold-start} problem ubiquitous in the literature.

\begin{figure}[!hbt]
    \centering
    \includegraphics[width=0.85\linewidth]{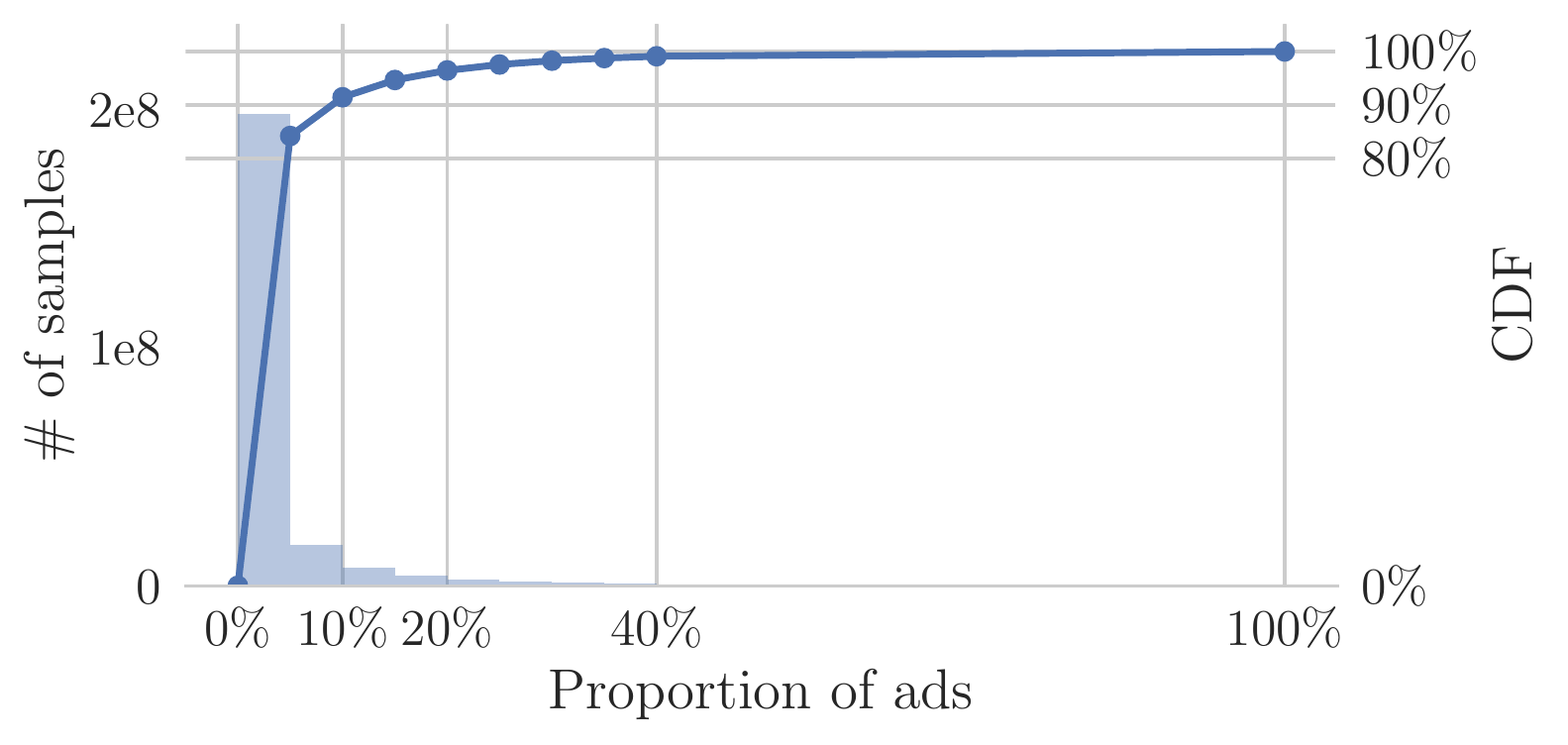}
\caption{Histogram of the number of samples over different proportions of ads of the KDD Cup 2012 search ads dataset.}
\label{fig:hist}\end{figure}

\begin{figure*}[hbt!]
    \centering
    \begin{minipage}{0.5\linewidth} \centering
    \includegraphics[width=0.85\linewidth]{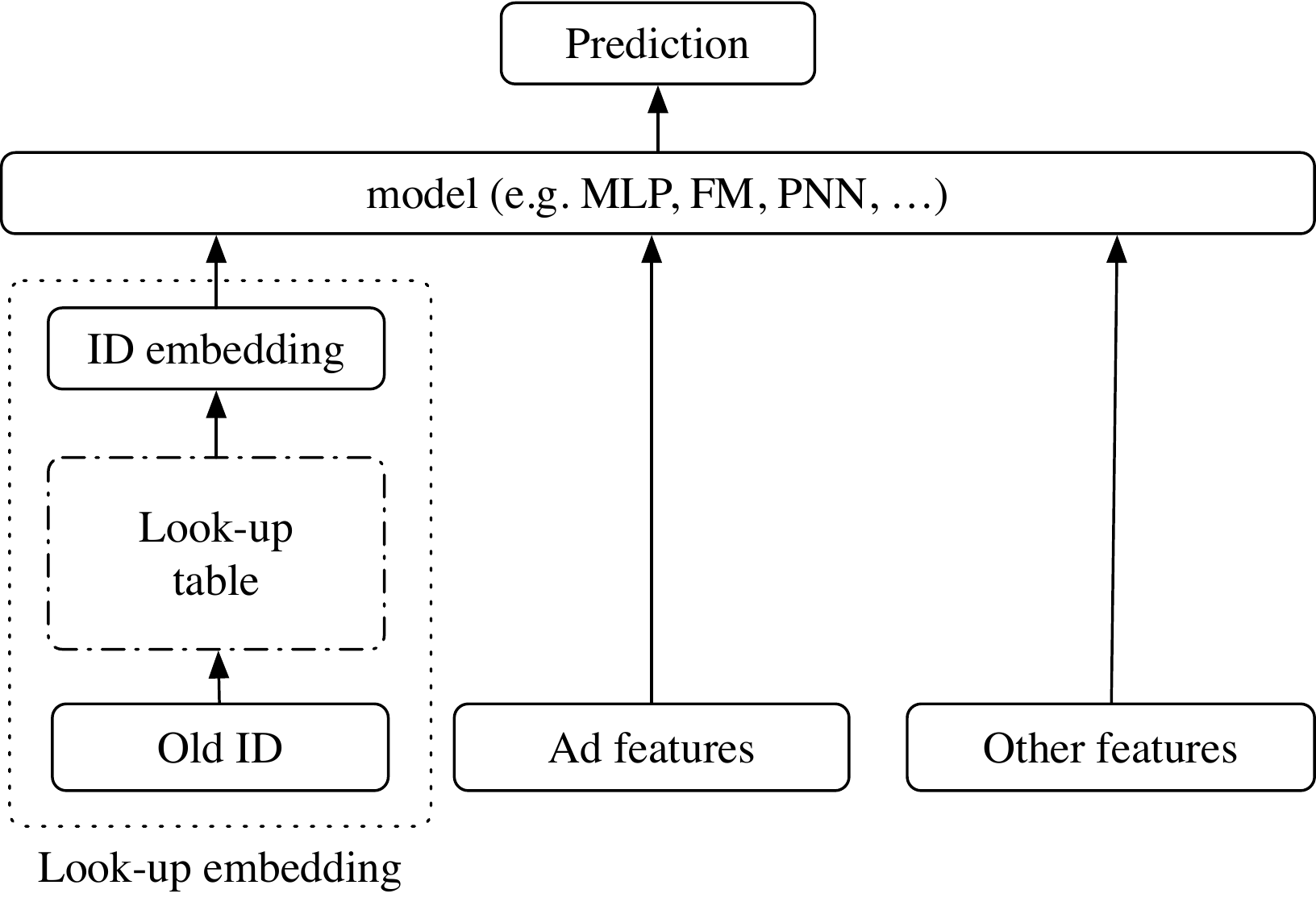}
    \\(a) Warm-start: look-up embeddings for old ads
    \end{minipage}
    \begin{minipage}{0.5\linewidth}
    \centering
    \includegraphics[width=0.85\linewidth]{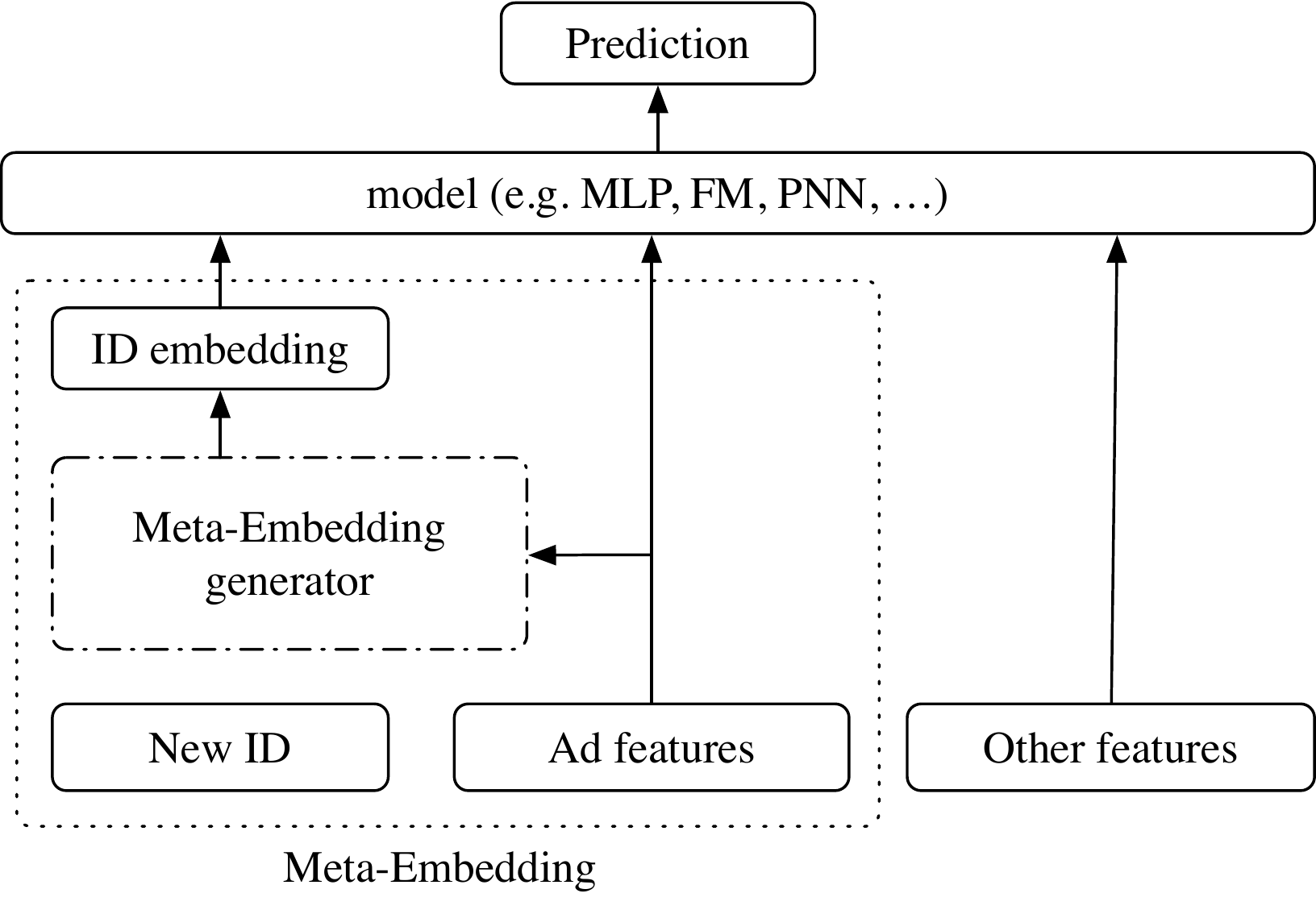}
    \\(b) Cold-start: Meta-Embedding for new ads
    \end{minipage}
\caption{Comparison: Look-up Embedding for old IDs and Meta-Embedding for new IDs. For old ads, we look up the embedding from the look-up table trained with labeled samples previously. For new IDs in the cold-start phase, we use Meta-Embedding to generate an initial embedding by feeding the ad features.}
\label{fig:lookup-and-meta}\end{figure*}

The cold-start problem has become a crucial obstacle for online advertising. For example, Figure \ref{fig:hist} shows that in the KDD Cup 2012 dataset\footnote{http://www.kddcup2012.org/c/kddcup2012-track2} of search ads, $5\%$ of ads accounted for over $80\%$ of samples, while the rest of $95\%$ small-sized ads had a very small amount of data. Therefore, being good at cold-start can not only benefit the revenue but also improve the satisfaction of small advertisers.

In light of these observations, we aim to develop a new method that can warm up cold advertisements. Our high-level idea is to learn better initial embeddings for new ad IDs that can 1) yield acceptable predictions for cold-start ads and 2) to speed up the model fitting after a small number of examples are obtained. We propose a meta-learning approach that learns how to learn such initial embeddings, coined {\em Meta-Embedding} in this paper. The main idea of the proposed method includes leveraging a parameterized function of ad features as the ID embedding generator, and a two-phase simulation over old IDs to train the generator by gradient-based meta-learning to improve both cold-start and warm-up performances.

To start with, we list two important desiderata that we pursue:
\begin{enumerate}
\item \textbf{Better at cold-start}:
When making predictions for a new ad with no labeled data, one should make predictions with a smaller loss;
\item \textbf{Faster at warming-up}:
After observing a small number of labeled samples, one should speed up the model fitting so as to reduce the prediction losses for subsequent predictions.
\end{enumerate}

In order to achieve these two desiderata, we design a two-phase simulation over the ``large'' ads in hand. The simulation consists of a cold-start phase and a warm-up phase. At the cold-start phase, we need to assign an initial embedding for the ID with no labeled data. At the warm-up phase when we have access to a minimal number of labeled examples, we update the embedding accordingly to simulate the model fitting procedure. In this way, we can learn how to learn.

With the two-phase simulation, we propose a meta-learning algorithm to train the Meta-Embedding generator. The essence of our method is to recast CTR prediction as a meta-learning problem, in which learning each ad is viewed as a task. We propose a gradient-based training algorithm with the advantage of Model-Agnostic Meta-Learning (MAML)~\cite{finn2017model}. MAML is successful for fast-adaptation in many areas, but it trains one model per task, so it cannot be used for CTR predictions if there are millions of tasks (ads). To this end, we generalize MAML into a content-based embedding generator. We construct our unified optimization objective that balances both cold-start and warm-up performance. Then the generated embedding cannot only yield acceptable predictions with no labeled data, but also adapt fast when a minimal amount of data is available. Lastly, our method is easy to implement and can be applied either offline or online with static or streaming data. It can also be used to cold-start other ID features, e.g., the user ID and the advertiser ID.

\begin{figure*}[hbt!]
    \centering
    \includegraphics[width=\linewidth]{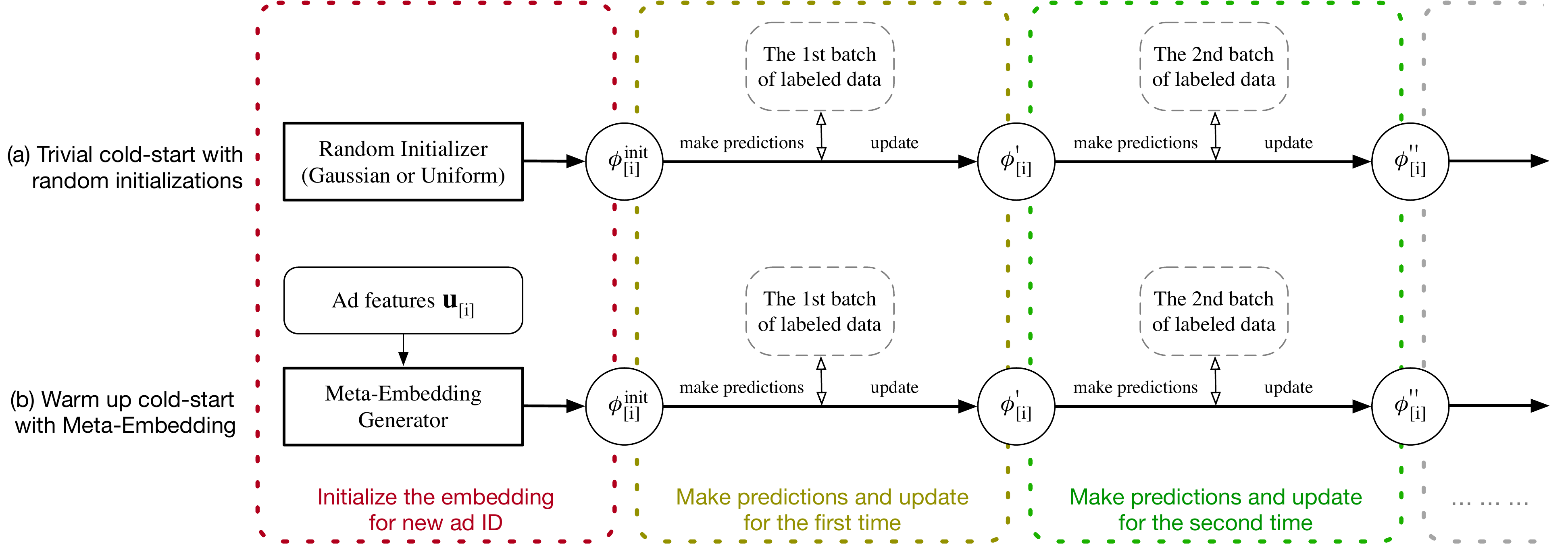}
\caption{Meta-Embedding is easy to deploy. It is compatible with the standard updating scheme. From left to right, it first initializes the embedding before observing any labeled data. The second block is when we observe the data, receive the prediction loss, and update the embedding for the first time. The third block is when we apply the adapted embedding and receive losses again. We are motivated to find a good embedding generator that can reduce all these losses, to warm up cold-start.}
\end{figure*}

Note that all the models and experiments in this paper are under the general online supervised learning framework where the system passively collects data. So we do not address the trade-off of exploration and exploitation \cite{li2010contextual,nguyen2014cold,tang2015personalized,shah2017practical,pan2018policy,pan2018pome}, nor do we design interviews as in active learning \cite{park2006naive,zhou2011functional,golbandi2011adaptive,harpale2008personalized}. But our methodology can be easily extended to these cases.

To summarize, the contributions of this paper are as follows:
\begin{itemize}
\item We propose Meta-Embedding that learns how to learn embeddings for new ad IDs to address the cold-start problem. It generates initial embeddings for new ad IDs with the ad contents and attributes as inputs. To the best of our knowledge, it is the first work that aims to improve the performance of various state-of-the-art CTR prediction methods for both cold-start and warm-up phases for new ads under a general supervised learning framework.
\item We propose a simple yet powerful algorithm to train the Meta-Embedding generator using gradient-based meta-learning by making use of second derivatives when back-propagating.
\item The proposed method is easy to implement in the online cold-start setting. Once the embedding generator is trained, it can take the place of trivial random initializer for new ID embeddings so as to warm up cold-start for the new ad.
\item We verify the proposed methods on three large-size real-world datasets. Experimental results show that six existing state-of-the-art CTR prediction methods can all be drastically improved when leveraging Meta-Embedding for both cold-start new ads and warm-up small-sized ads.
\end{itemize}

\section{Background and Formulations}

CTR prediction is a supervised binary classification task. Each instance $(\mathbf{x}, y)$ consists of the input features $\mathbf{x}$ and the binary label $y$. A machine learning system should approximate the probability $p=\Pr(y=1\mid\mathbf{x})$ for the instance with input $\mathbf{x}$.

In most cases, the inputs $\mathbf{x}$ can be splitted into three parts, i.e., $\mathbf{x} = (\id, \uid, \bv)$, including

\begin{enumerate}
\item $\id$, the ad identifier (ID), a positive integer to identify each ad;
\item $\uid$, the features and attributes of the specific ad $\id$, may have multiple fields;
\item $\bv$, other features which do not necessarily relate to the ad, may have multiple fields, such as the user features and contextual information.
\end{enumerate}
So we predict $p$ by a discriminative function of these three parts,
\begin{equation}
\hat p = f(\id, \uid, \bv).
\label{pure-f}\end{equation}
Because the ID is an index, it has to be encoded into real-value to apply gradient-based optimization methods. One-hot encoding is a basic tool to encode it into a sparse binary vector, where all components are $0$ except for the $\id^{\textrm{th}}$ component is $1$. For example,
\begin{equation}
\id=3 \xrightarrow{\textrm{one-hot}} \mathbf{e}_{\id} = (0,0,1,0,\dots)^T \in \mathbb{R}^n,
\end{equation}
where $n$ is the number of IDs in all. If followed up by a matrix multiplication, we will have a low-dimensional representation of the ID, i.e., given a dense matrix $\Phi\in \mathbb{R}^{n\times m}$, then
\begin{equation}
\phi_{[\id]}=\mathbf{e}_{\id}^T \Phi \in \mathbb{R}^m
\label{phi_id_ohe}\end{equation}
is the resulted $m$-dimentional dense real-valued vector for ID $\id$. In this way, the ID is turned into a low-dimensional dense vector.

However, due to the fact that there are often millions of IDs, the one-hot vector can be extremely large. So nowadays it is more usual to use the look-up embedding as a mathematically equivalent alternative. For ID $\id$, it directly looks up the $\id^{\textrm{th}}$ row from $\Phi$,
\begin{equation}
\id=3 \xrightarrow{\textrm{look-up the }\id^{\textrm{th}}\textrm{ row from } \Phi} \phi_{[\id]} \in \mathbb{R}^m
\label{phi_id_lookup}\end{equation}
where the resulted $\phi_{[\id]}$ is the same as in (\ref{phi_id_ohe}). The dense matrix $\Phi$ is often referred to as the embedding matrix, or the look-up table.


Given $\Phi$ the embedding matrix and $\phi_{[\id]}$ the embedding for ID $\id$, we can get a parameterized function as the discriminative model,
\begin{equation}
\hat p = f_\btheta(\phi_{[\id]}, \uid, \bv).
\label{para-f}\end{equation}
For readability, we refer to the function $f_\btheta$ as the \textit{base model} throughout this paper, and let $\btheta$ denote its parameters. Figure \ref{fig:lookup-and-meta}(a) shows the basic structure of such a model.

The Log-loss is often used as the optimization target, i.e.
\begin{equation}
l(\btheta, \Phi) = -y \log \hat p- (1-y)\log(1-\hat p).
\end{equation}
In offline training where a number of samples for each ad ID are available, $\btheta$ and $\Phi$ are updated simultaneously to minimize the Log-loss by Stochastic Gradient Descent (SGD).

What if there comes an ID $\newid$ that the system has never seen before? It means that we have never got labeled data for the ad, so the corresponding row in the embedding matrix remains an initial state, for example, random numbers around zero. Therefore the testing accuracy would be low. It is known as the ad cold-start problem.

To address this issue, we aim to design an embedding generator. The structure is shown in Figure \ref{fig:lookup-and-meta}(b).
Given an unseen ad $\newid$, the generator inputs the features $\unewid$ of the ad, and can somehow output a ``good'' initial embedding as a function of these features, $$\phiiinew=h_{\bw}(\unewid).$$

Now that we explained why we want to design the embedding generator to address the cold-start problem. The problem is, how to train such an embedding generator? For what objective do we update its parameters?
To accomplish our final goal of warming up cold-start, we propose a meta-learning based approach which learns how to learn new ad ID embeddings. Specifically, the parameters $\bw$ in the embedding generator are trained by making use of previously learned ads through gradient-based meta-learning, which will be detailed in Section~\ref{sec:model}.

\section{Learning to learn the ID embeddings for new ads}\label{sec:model}

In this section, we propose Meta-Embedding, a meta-learning based method for learning to learn the ID embeddings for new ads, to solve the ad cold-start problem.
First, we recast the CTR prediction problem as meta-learning.
Then, we propose the framework of learning to learn ID embeddings and introduce our meta-learner induced by both the pre-trained model and a Meta-Embedding generator.
Finally, we detail the model architectures and hyper-parameters.
\subsection{Recasting CTR prediction as Meta-Learning}
\label{ME-recasting}

Recall that the final goal for CTR predictions as shown in (\ref{pure-f}) is to learn a predictive function $f(\cdot)$ with inputs of an ID and two sets of features. Only after the ID is transformed into a real-valued vector can we begin to learn the parameterized base model $f_{\btheta}(\phi_{[\id]},\cdot,\cdot)$ as in (\ref{para-f}). Therefore, for a given ad, the embedding of ID implicitly determines the hidden structure of the model.

To look at it from a meta-learning perspective, we introduce a new notation to write the predictive model for a fixed ID $\id$ as
\begin{equation}
\hat p=g_{[\id]}(\uid, \bv)=f_{\btheta}(\phi_{\id}, \uid, \bv).
\end{equation}
Note that $g_{[\id]}(\cdot,\cdot)$ is exactly the same function as $f_{\btheta}(\phi_{[\id]}, \cdot,\cdot)$, whose parameters are $\btheta$ and $\phi_{[\id]}$.

In this way, we can see CTR prediction as an instance of meta-learning by regarding the learning problem w.r.t each ad ID as a distinguished task. For IDs $\id=1,2,\cdots$, the tasks $t_1,t_2,\cdots$ are to learn the task-specific models $g_{[1]},g_{[2]},\cdots$ respectively. They share the same set of parameters $\btheta$ from the base model, and meanwhile maintain their task-specific parameters $\phi_{[1]}, \phi_{[2]}, \cdots$.

Consider that we have access to prior tasks $t_\id$ with IDs $\id \in \mathcal{I}$, the set of all known IDs, as well as a number of training samples for each task. The original (pre-)training on this data gives us a set of well-learned shared parameters $\btheta$ and task-specific parameters $\phi_{[\id]}$ for all prior IDs $\id \in \mathcal{I}$. However, for a new ID $\newid \not \in \mathcal{I}$, we do not know $\phi_{[\newid]}$. So we desire to learn how to learn $\phi_{[\newid]}$ with the prior experience on those old IDs. This is how we see the cold-start problem of CTR prediction from a meta-learning point of view.
\subsection{Meta-Embedding}
\label{ME-l2l}
In this section, we put forward Meta-Embedding that captures the skills of learning ID embeddings for new ads via meta-learning.

In the previous section, we introduced the shared parameters $\btheta$ and task-specific parameters $\phi_{\id}$ for old items $\id\in\mathcal{I}$. As far as $\btheta$ is usually trained previously with an extremely large amount of historical data, we are confident about its effectiveness. So, when training the Meta-Embedding, we freeze $\btheta$ and do no update it during the whole process. The only thing matters for the cold-start problem in this paper is how to learn the embeddings for new IDs.

Recall that the task-specific parameter $\phi_{[\id]}$ is unknown for any unseen ID. So we need to use a shared functional embedding generator to take its place. For a new ad with ID $\newid$, we let
\begin{equation}
\phiiinew=h_\bw(\unewid),
\end{equation}
as the generated initial embedding for simplicity of notations. Then the model induced by the generated embedding is
\begin{equation}
g_{\meta}(\unewid, \bv) = f_{\btheta}\big(\phiiinew, \unewid, \bv\big).
\end{equation}So here $g_\meta(\cdot, \cdot)$ is a model (a meta-learner) that inputs the features and outputs the predictions, without involving the embedding matrix. The trainable parameter for it is the meta-parameter $\bw$ from $h_\bw(\cdot)$.

 Now we describe the detailed procedure to simulate cold-start with old IDs as if they were new.
Consider that for each task $t_\id$ w.r.t an old ID $\id$, we have already got training samples $\mathcal{D}_{[\id]}=\big\{(\uid, \bv_j)\big\}_{j=1}^{N_\id}$, where $N_\id$ is the number of samples for the given ID.

To begin with, we randomly select two disjoint minibatches of labeled data, $\mathcal{D}_{[\id]}^{a}$ and $\mathcal{D}_{[\id]}^{b}$, each with $K$ samples. It is assumed that the minibatch size is relatively small, i.e., $K << N_\id/2$.

\begin{figure*}[hbt!]
    \centering
    \includegraphics[width=0.95\linewidth]{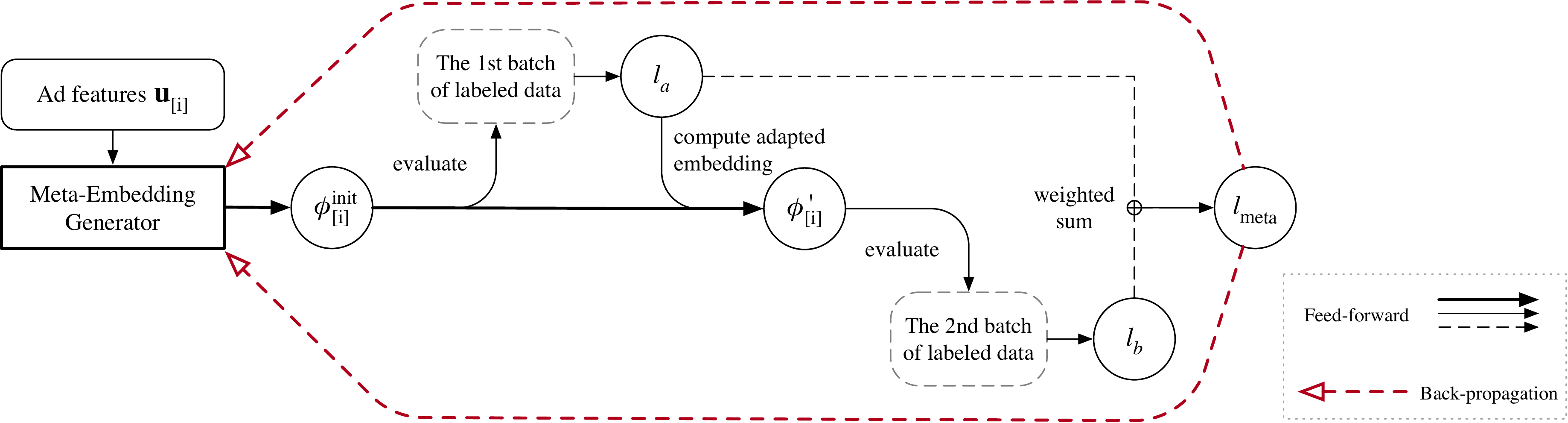}
\caption{Demonstration of how to train the parameters of the Meta-Embedding generator, corresponding to the inner for-loop of Alg.\ref{alg:one}. The use of second derivatives make our method powerful for warm up cold-start. The training for Meta-Embedding can either use offline or online data, or both.}
\label{fig:train}\end{figure*}
\subsubsection{Cold-start phase}
We first make predictions using $g_\meta(\cdot,\cdot)$ on the first minibatch $\mathcal{D}_{[\id]}^{a}$, as
\begin{equation}
\hat p_{aj} = g_\meta(\uid,\bv_{aj}) = f_\btheta(\phiii, \uid, \bv_{aj})
\end{equation}
where the subscript `$aj$' is for the $j^\textrm{th}$ sample from batch $\mathcal{D}_{[\id]}^{a}$. Then we calculate the average loss over these samples
\begin{equation}
l_a = \frac{1}{K} \sum_{j=1}^K\bigg[ -y_{aj} \log \hat p_{aj}- (1-y_{aj})\log(1-\hat p_{aj})\bigg].
\end{equation}
By far, we have done with the cold-start phase: we generated the embedding $\phiii$ by the generator $h_\bw(\cdot)$, and evaluated it on the first batch of data to get a loss $l_a$.
\subsubsection{Warm-up phase}
Next, we simulate the learning process for the warm-up phase with the second batch of data $\mathcal{D}_{[\id]}^{b}$.

By computing the gradient of $l_a$ w.r.t the initial embedding $\phiii$ and take a step of gradient descent, we get a new adapted embedding
\begin{equation}
\phiprime = \phiii - a\,\frac{\partial l_a}{\partial \phiii}
\end{equation}
where  $a>0$ is the step size of gradient descent.

Now that we have a new embedding which is trained with a minimal amount of data, we can test it on the second batch of data. Similar to the previous part, we make predictions
\begin{equation}
\hat p_{bj} = g'_{[\id]}(\uid,\bv_{bj}) = f_\btheta(\phiprime, \uid, \bv_{bj})
\end{equation}
and compute the average loss
\begin{equation}
l_b = \frac{1}{K} \sum_{j=1}^K\bigg[ -y_{bj} \log \hat p_{bj}- (1-y_{bj})\log(1-\hat p_{bj})\bigg].
\end{equation}

\subsubsection{A unified optimization objective}\label{ME-obj}
We suggest to evaluate the goodness of the initial embeddings from two aspects:
\begin{enumerate}
\item The error of CTR predictions for the new ad should be small;
\item After a small number of labeled examples are collected, a few gradient updates should lead to fast learning.
\end{enumerate}

Surprisingly, we find that the two losses $l_a$ and $l_b$ can perfectly suit these two aspects respectively. On the first batch, since we make predictions with generated initial embeddings, $l_a$ is a natural metric to evaluate the generator in the cold-start phase. For the second batch, as the embeddings have been updated once, it is straight-forward that $l_b$ can evaluate the sample efficiency in the warm-up phase.

To unify these two losses, we propose our final loss function for Meta-Embedding as a weighted sum of $l_a$ and $l_b$,
\begin{equation}
l_\meta = \alpha l_a + (1-\alpha) l_b,
\label{ME-loss}\end{equation}
where $\alpha \in [0,1]$ is a coefficient to balance the two phases.

As $l_\meta$ is a function of the initial embedding, we can back-prop the gradient w.r.t the meta-parameter $\bw$ by the chain rule:
\begin{equation}
\frac{\partial l_\meta}{\partial \bw} = \frac{\partial l_\meta}{\partial \phiii}\frac{\partial \phiii}{\partial \bw} = \frac{\partial l_\meta}{\partial \phiii}\frac{\partial h_\bw}{\partial \bw},
\end{equation}
where
\begin{align}
\frac{\partial l_\meta}{\partial \phiii}
&= \alpha \frac{\partial l_a}{\partial \phiii} + (1-\alpha) \frac{\partial l_b}{\partial \phiprime} - a (1-\alpha) \frac{\partial l_b}{\partial \phiprime}\frac{\partial^2 l_a}{\partial {\phiii}^2}.
\end{align}
Although it involves second derivatives, i.e., a Hessian-vector, it can be efficiently implemented by existing deep learning libraries that allows automatic differentiation, such as TensorFlow \cite{abadi2016tensorflow}.

Finally, we come to our training algorithm, which can update the meta-parameters by stochastic gradient descent in a mini-batch manner, see Alg.\ref{alg:one}. A demonstration of training procedure w.r.t each single ID (the inner for-loop of Alg.\ref{alg:one}) is also shown in Figure \ref{fig:train}.

Note that Meta-Embedding not only can be trained with offline data set, but also can be trained online with minor modifications by using the emerging new IDs as the training examples.
\begin{algorithm}
\caption{Train Meta-Embedding by SGD}\label{alg:one}
\begin{algorithmic}[1]
\Require $f_\btheta$: the pre-trained base model.
\Require $\mathcal{I}$: the set of all existing IDs.
\Require $\alpha$: hyper-parameter, the coefficient for meta-loss.
\Require $a,b$: step sizes.
\State Randomly initialize $\bw$
\While{not done}
\State Randomly sample $n$ IDs $\{\id_1, \dots, \id_n\}$ from $\mathcal{I}$
    \For{$\id\in \{\id_1, \dots, \id_n\}$}
    \State Generate the initial embedding: $\phiii=h_\bw(\uid)$
    \State Sample mini-batches $\mathcal{D}^a_{[\id]}$ and $\mathcal{D}^b_{[\id]}$ each with $K$ samples
    \State Evaluate loss $l_a(\phiii)$ on $\mathcal{D}^a_{[\id]}$
    \State Compute adapted embedding:
    $\phiprime = \phiii - a\,\frac{\partial l_a(\phiii)}{\partial \phiii}$
    \State Evaluate loss $l_b(\phiprime)$ on $\mathcal{D}^b_{[\id]}$
    \State Compute loss: $l_{\meta,\id}=\alpha l_a(\phiii) + (1-\alpha)l_b(\phiprime)$
    \EndFor
\State Update $\bw\leftarrow \bw-b \sum_{\id\in \{\id_1, \dots, \id_n\}}\frac{\partial l_{\meta,\id}}{\partial \bw}$
\EndWhile
\end{algorithmic}
\end{algorithm}

\subsection{Architecture and hyper-parameter}
Finally, we discuss how to choose the architecture of the embedding generator along with the hyper-parameters.

\begin{figure}[hbt!]
    \centering
    \includegraphics[width=\linewidth]{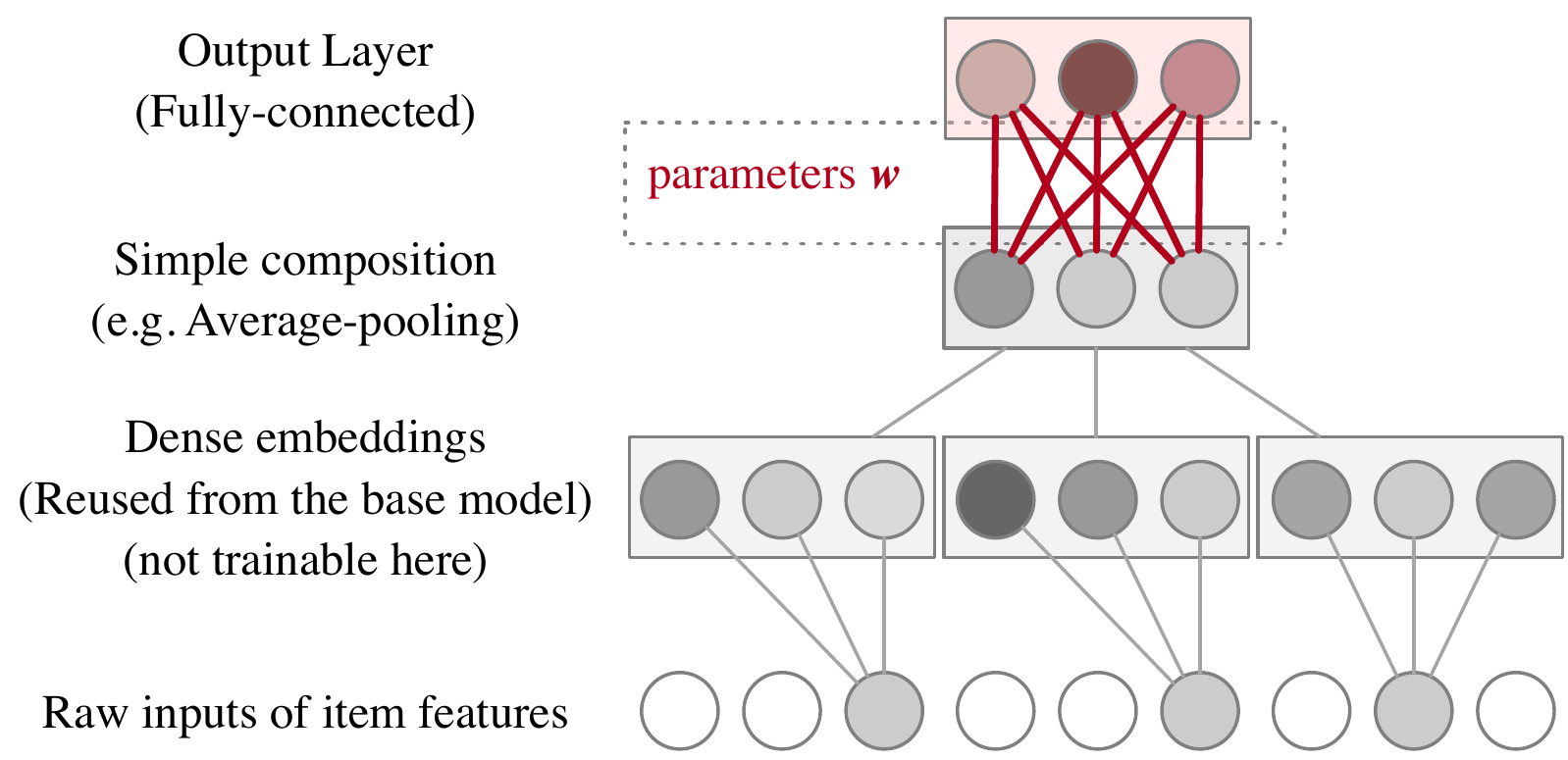}
\caption{An example of the Meta-Embedding generator, with frozen reused embedding layers reused from the base model and lightweight compositional layers.}
\label{fig:net}\end{figure}
\subsubsection{Architectures of $h_\bw(\cdot)$}
In principle, we want the embedding generator to be simple yet strong. So we suggest using the neural network as the parameterized function. Any common net architecture can be used for Meta-Embedding.

Here we would like to show an instance of the generator whose effectiveness is tested in our experiments, see Figure \ref{fig:net}. It is simple and lightweight, thus can be a basic choice if anyone would like to start using Meta-Embedding.

The input of the generator network is the ad feature $\uid$. However, we do not need to train the generator from scratch. Recall that, in the base model, there is already embedding layers for ad features, so we can directly reuse these layers. We suggest freezing the weights of the reused layers so as to reduce the number of trainable parameters. The embeddings from different fields can be aggregated by average-pooling, max-pooling, or concatenating. Finally, we use a dense layer to get the outputs, which is the only trainable part of the generator. We can have numerically stable outputs by using three tricks on the final layer: 1) use the tanh activation, 2) do not add the bias term, and 3) use L2 regularization to penalize the weights.

\subsubsection{The hyper-parameter $\alpha$}
The coefficient $\alpha\in[0,1]$ balances the cold-start loss and the warm-up loss. Since the warm-up phase usually involves more time steps then the cold-start phase in practice, we suggest to set it to a small value so that the meta-learner will pay more attention to the warm-up phase to achieve fast adaptation. Empirically, we found that our algorithm is robust to the hyper-parameter. We simply set $\alpha$ to $0.1$ for all our experiments.
\section{Experiments}

\subsection{Datasets}
We evaluate Meta-Embedding on three datasets:

\textbf{MovieLens-1M\footnote{http://www.grouplens.org/datasets/movielens/}:} One of the most well-known benchmark data. The data consists of 1 million movie ranking instances over thousands of movies and users. Each movie can be seen as an ad in our paper, with features including its title, year of release, and genres. Titles and genres are lists of tokens. Other features include the user's ID, age, gender, and occupation. We first binarize ratings to simulate the CTR prediction task, which is common to test binary classification methods \cite{kozma2009binary,volkovs2015effective}. The ratings at least 4 are turned into 1 and the others are turned into 0. Although this dataset is not a dataset for CTR prediction, its small size enables us to easily verify the effectiveness of our proposed method.

\textbf{Tencent CVR prediction dataset for App recommendation:} The public dataset for the Tencent Social Ads competition in 2018\footnote{http://algo.qq.com/} with over 50 million clicks. Each instance is made up by an ad, a user and a binary label (conversion). Each ad has 6 categorical attributes: advertiser ID, ad category, campaign ID, app ID, app size, and app type. Other features are user attributes, including the user's gender, age, occupation, consumption ability, education level, and city.

\textbf{KDD Cup 2012 CTR prediction dataset for search ads}\footnote{https://www.kaggle.com/c/kddcup2012-track2}: The dataset contains around 200 million instances derived from session logs of the Tencent proprietary search engine, \textit{soso.com}. Each ad has three features: keywords, title, and description. These features are lists of anonymized tokens hashed from natural language. Other features consist of the query (also a list of tokens), two session features and the user's gender and age.

\subsection{Base models}
Because our Meta-Embedding is model-agnostic, it can be applied upon various existing models that in the Embedding \& MLP paradigm. To show the effectiveness of our method, we conduct experiments upon the following representative models:
\begin{itemize}
\item \textbf{FM}: the 2-way Factorization Machine \cite{rendle2010factorization}. Originally it uses one-hot encodings and matrix multiplications to get embedding vectors, while in our implementation we directly use look-up embeddings. For efficiency, we use the same embedding vectors for the first- and the second-order components.
\item \textbf{Wide \& Deep}: proposed by Google in \cite{cheng2016wide} which models both low- and high-order feature interactions. The wide component is a Logistic Regression that takes one-hot vectors as inputs. The deep component has embedding layers and three dense hidden layers with ReLU activations.
\item \textbf{PNNs}: Product-based Neural Networks~\cite{qu2016product}. It first feeds the dense embeddings into a dense layer and a product layer, then concatenates them together and uses another two dense layers to get the prediction. As suggested by their paper, we use three variants: \textbf{IPNN}, \textbf{OPNN}, and \textbf{PNN*}. IPNN is the PNN with an inner product layer, OPNN is the PNN with an outer product layer, and PNN* has both inner and outer products.
\item \textbf{DeepFM} \cite{guo2017deepfm}: a recent state-of-the-art method that learns both low- and high-level interactions between fields. It feeds dense embeddings into an FM and a deep component, and then concatenates their outputs and gets the final prediction by a dense layer. For the deep component, we use three ReLU layers as suggested in \cite{guo2017deepfm}.
\end{itemize}

These base models share the common Embedding \& MLP structure. The dimensionality of embedding vectors of each input field is fixed to 256 for all our experiments. For natural language features in MovieLens and the KDD Cup dataset, we first embed each token (word) into a 256-dimensional word-embedding, then use AveragePooling to get the field-(sentence-)level representation. In other words, after the embedding layer, every field is embedded into a 256-dimensional vector respectively.

\subsection{Experiment Set-Up}

\begin{table*}
\caption{The statistics for data splitting}
\begin{center}
\small
\begin{tabular}{|l||r||r|r|r||r|r|r|}
\hline
\multirow{3}{*}{Dataset} &
\multirow{3}{*}{Minibatch size $K$} &
\multicolumn{3}{ c|| }{Old ads} &
\multicolumn{3}{ c| }{New ads} \\
\cline{3-8}
 & & \# of IDs & \# of samples & \# of samples used to &
\# of IDs & \# of samples & \# of samples in \\
& & & & train Meta-Embedding &
 & & the hold-out set\\
\hline
MovieLens-1M & 20 & 1127 & 0.76 M & 0.09 M & 1058 & 0.19 M & 0.12 M\\
\hline
Tencent CVR data & 200 & 572 & 49.33 M & 0.45 M & 443 & 5.00 M & 4.74 M\\
\hline
KDD Cup 2012 & 200 & 6534 & 148.55 M & 5.22 M & 9299 & 28.71 M & 23.13 M\\
\hline
\end{tabular}
\end{center}
\label{Table:data-split}\end{table*}
\begin{figure*}[hbt!]
\centering
\includegraphics[width=\linewidth]{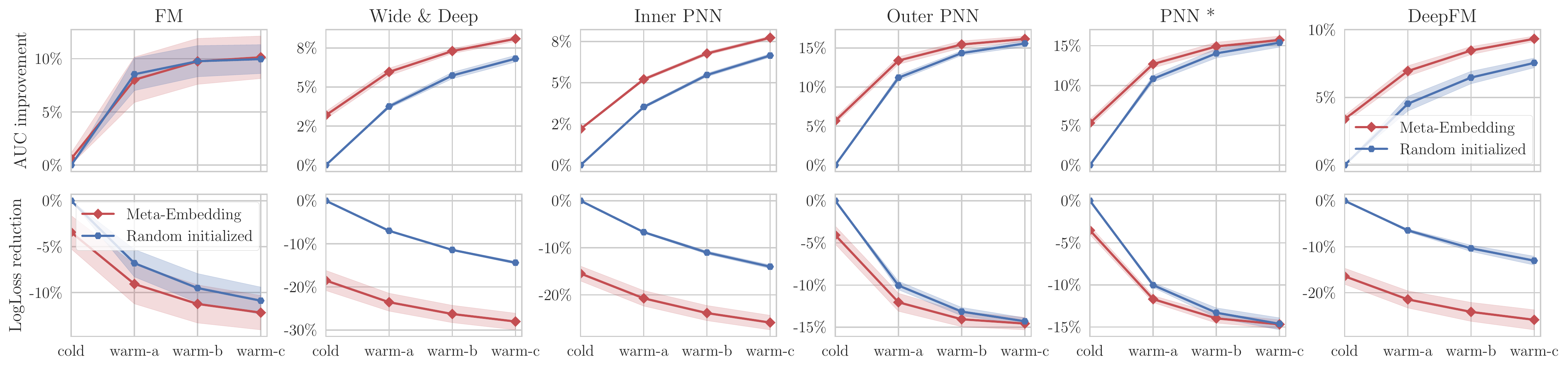}
\caption{Cold-start performance in percentage on MovieLens-1M, over six popular base models. The solid lines are averaged scores and the bands are standard deviations over three runs.}
\label{fig:movielens}\end{figure*}

\subsubsection{Dataset splits}
To evaluate the performance of cold-start advertising in the two phases, we conduct the experiments by splitting the datasets. First, we group the advertisements by their sizes:
\begin{itemize}
\item Old ads: the ads whose number of labeled instances is larger than a threshold $N$. Since the volumes of the three datasets are different, we used $N$ of $300$, $20000$, and $5000$ for MovieLens-1M, Tencent CVR data, and KDD Cup data, respectively.
\item New ads: the ads whose number of labeled instances is less than $N$, but larger than $N_\textrm{min}>3\times K$, where $K$ is the mini-batch size mentioned in Section 3. The reason for setting the minimum number of samples is to guarantee that after cutting $3\times K$ samples for the warm-up phase, there still remains a number of samples for testing. $N_\textrm{min}$ is set to $80$, $2000$, and $2000$ for the three datasets, respectively.
\end{itemize}

The summary of the data split can be seen in Table \ref{Table:data-split}. We set the thresholds to make the number of new ads be close to the number of old ads, although the data size of new ads is much smaller.
\subsubsection{Experiment pipeline}

To start with, we use the old ads to pre-train the base models. Then we use them to conduct offline training for Meta-Embedding. The number of samples used to train Meta-Embedding can be also found in Table \ref{Table:data-split}. After training, it is tested over new ads.
For each new ad, we train on 3 mini-batches one by one as if they were the warming-up data, named batch-a, -b, and -c, each with $K$ instances. Other instances for new ads are hold-out for testing.

The experiments are done with the following steps:
\begin{enumerate}
\item[0.] Pre-train the base model with the data of old ads (1 epoch).
\item[1.] Train the Meta-Embedding with the training data (2 epochs).
\item[2.] Generate initial embeddings of new ad IDs with (random-initialization or Meta-Embedding).
\item[3.] Evaluate cold-start performance over the hold-out set;
\item[4.] Update the embeddings of new ad IDs with batch-a and compute evaluation metrics on the hold-out set;
\item[5.] Update the embeddings of new ad IDs with batch-b and compute evaluation metrics on the hold-out set;
\item[6.] Update the embeddings of new ad IDs with batch-c and compute evaluation metrics on the hold-out set;
\end{enumerate}

The training effort for Meta-Embedding is very small compared to the pre-train step. We train it with $2\times K$ samples per epoch/ad, so the number of samples involved is only (\# of old IDs) $\times K \times 2 \times 2$. The actual number could be found in Table \ref{Table:data-split}.

\subsubsection{Evaluation metrics}
We take two common metrics to evaluate the results, the Log-loss and the AUC score (Area Under Receiver Operator Characteristic Curve). All the metrics are evaluated only on the hold-out set.

To increase readability, we will show the relative improvements in percentage over the baseline cold-start score:
\begin{align*}
\textrm{LogLoss\_precentage}&=\bigg(\frac{\textbf{LogLoss}}{\textrm{LogLoss}_{\textrm{base-cold}}}-1\bigg)\times 100\%\\
\textrm{AUC\_precentage}&=\bigg(\frac{\textbf{AUC}}{\textrm{AUC}_{\textrm{base-cold}}}-1\bigg)\times 100\%
\end{align*}
where ``base-cold'' refer to cold-start scores produced by the base model with random initialized embeddings for unseen IDs. Thus, for LogLoss, it is better if the percentage is more negative. For AUC, the percentage is positive, the larger the better. Note that this ``base-cold'' is a strong baseline because it has all the warm contents to use, and is the commonly used in industry. 

\subsection{Experiment results}

\begin{table*}[!hbt]
\caption{Experimental results: Average performances on tested datasets, over six base models, three runs for each.} 
\begin{center}
\small
\begin{tabular}{|l||l||r|r||r|r||r|r||r|r|}
\hline
\multirow{2}{*}{Dataset} &
\multirow{2}{*}{Metrics} &  
\multicolumn{2}{ c|| }{Cold-Start phase} &
\multicolumn{2}{ c|| }{Warm-Up phase: a} & 
\multicolumn{2}{ c|| }{Warm-Up phase: b} & 
\multicolumn{2}{ c| }{Warm-Up phase: c} \\
\cline{3-10}
 & & baseline & Meta &
baseline & Meta & baseline & Meta & baseline & Meta\\
\hline
\multirow{2}{*}{MovieLens-1M} & AUC percentage & 0.0\% & +3.36\% & +7.02\% & +8.66\% & +9.25\% & +10.40\% & +10.27\% & +11.15\% \\
& Logloss percentage & 0.0\% & -10.23\% & -7.83\% & -16.42\% & -11.45\% & -18.93\% & -13.53\% & -20.20\% \\
\hline
\multirow{2}{*}{Tencent CVR data}& AUC percentage  & 0.0\% & +0.50\% & +0.60\% & +0.99\% & +1.16\% & +1.47\% & +1.57\% & +1.83\% \\
& LogLoss percentage  & 0.0\% & -0.19\% & -0.30\% & -0.44\% & -0.55\% & -0.65\% & -0.71\% & -0.79\% \\
\hline
KDD Cup 2012 & AUC percentage & 0.0\% & +0.73\% & +2.34\% & +2.76\% & +3.56\% & +3.87\% & +4.36\% & +4.61\% \\
Search Ads CTR& LogLoss percentage  & 0.0\% & -1.52\% & -1.97\% & -2.99\% & -3.06\% & -3.82\% & -3.79\% & -4.40\% \\
\hline
\end{tabular}
\end{center}
\label{Table:exp}\end{table*}
\begin{figure*}[hbt!]
\centering
\includegraphics[width=0.95\linewidth]{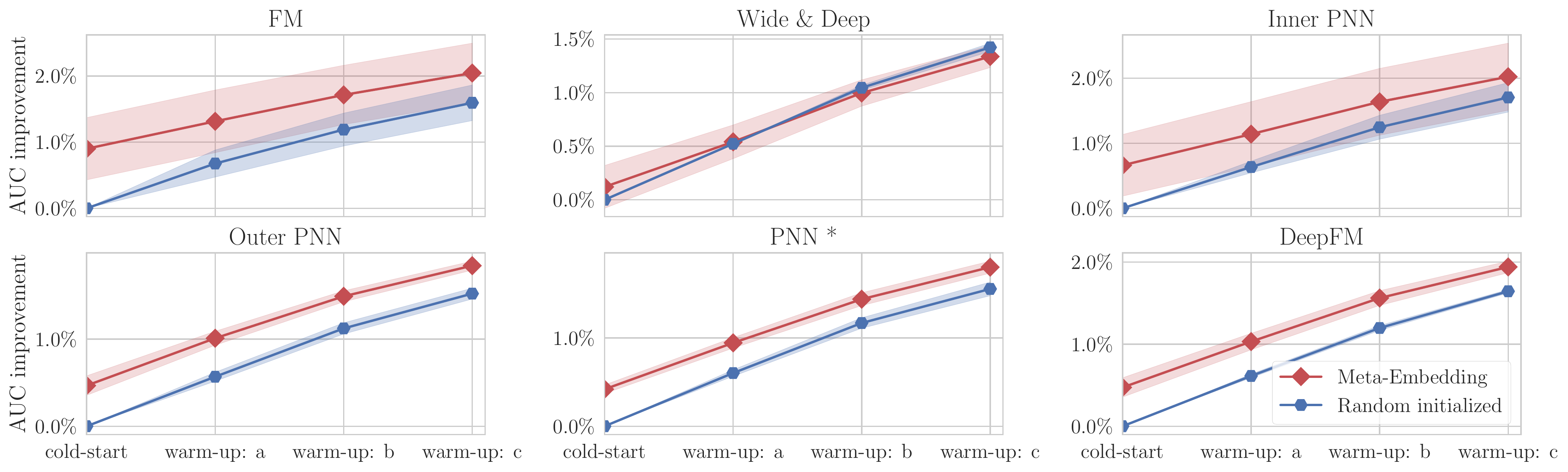}
\caption{AUC improvements in percentage on the Tencent CVR prediction data, over six base models.}
\label{fig:auc-tencent}\end{figure*}
\begin{figure*}[!hbt]
\centering
\includegraphics[width=0.95\linewidth]{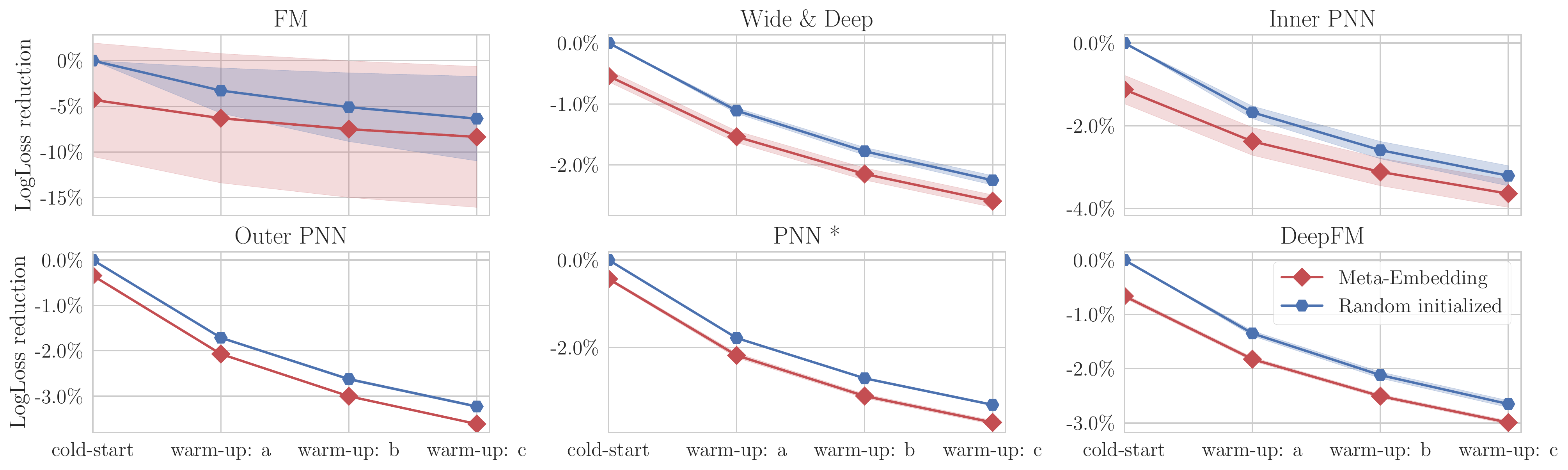}
\caption{LogLoss reductions in percentage on the KDD Cup 2012 CTR prediction dataset for search ads, over six models.}
\label{fig:logloss-kdd}\end{figure*}
For each experiment, we took three runs and reported the averaged performance.
The main experimental results are shown in Table \ref{Table:exp}.

We first see the results on MovieLens-1M. Figure~\ref{fig:movielens} exhibits the detailed performance of each tested model. We observed that when using Meta-Embedding, most models yielded better AUC and LogLoss results comparing to the random initialization baseline. Although the hyper-parameter of coefficient $\alpha$ was set to 0.1 so the major attention was on the warm-up loss, it still performed really well in the cold-start phase. For example, when using Wide\&Deep, IPNN and DeepFM as the base models, Meta-Embedding provided with a LogLoss reduction of over 15\% in the cold-start phase. After three times of warm-up updates, Meta-Embedding still outperformed the baseline, thus we are confident that Meta-Embedding increases the sample efficiency for new ads.

On the two larger datasets, Meta-Embedding also achieved considerable improvements as shown in Table \ref{Table:exp}. Due to the space limitation, we only show part of the detailed results: Figure \ref{fig:auc-tencent} shows the AUC improvements on the Tencent CVR prediction dataset, and Figure \ref{fig:logloss-kdd} shows the LogLoss on the KDD Cup search ad dataset.

An interesting finding is that the relative improvement was more significant on smaller datasets (MovieLens) than the larger ones. To explain, recall that the number of training examples in MovieLens is much smaller, so the shared pre-trained model parameters $\btheta$ may not be good enough. Therefore, in this case, there could be a larger improvement space for cold-start ads, for example, Meta-Embedding can sometimes double the improvement compared to the random initialization baseline. 

For most of the experiments, the scores were greatly improved by replacing random initialization with Meta-Embedding. But there were also some cases where the performances were mixed. For example, for MovieLens with FM as the base model, Meta-Embedding had similar AUC with the baseline. But the LogLoss of Meta-Embedding was smaller, so we can still be sure that Meta-Embedding can help FM. For Wide \& Deep in the Tencent CVR data (see Figure \ref{fig:auc-tencent}), the baseline tended to surpass Meta-Embedding after the third warm-up update. 
However, it only happens for the Tencent CVR data only. We think it was due to the ``Wide'' component which directly used raw inputs. We conjecture it mainly because this dataset had categorical inputs only. In the other two datasets, there are sequential inputs, so the ``Wide'' component can not dominate the performance.

Overall, we observe that Meta-Embedding is able to significantly improves CTR and CVR predictions for cold-start ads, and can work for almost all the tested base models and datasets in our comparisons.
\section{Related Work}
\subsubsection*{A. Methods addressing the cold-start problem} There are two types of methods. The first type actively solve cold-start by designing a decision making strategy, such as using contextual-bandits \cite{li2010contextual,caron2013mixing,nguyen2014cold,tang2015personalized,shah2017practical,pan2018policy}, or designing interviews to collect information for the cold item or user \cite{park2006naive,zhou2011functional,golbandi2011adaptive,harpale2008personalized}.

This paper belongs to the second type which treats cold-start within an online supervised learning framework. It is common to use side information for the cold-start phase, e.g., using user attributes \cite{seroussi2011personalised, zhang2014addressing, roy2016latent, volkovs2017dropoutnet}, item attributes \cite{zhang2014addressing, saveski2014item, schein2002methods, gu2010collaborative, mo2015image, volkovs2017dropoutnet, vartak2017meta}, or relational data \cite{yao2014dual, lin2013addressing, zhao2016connecting}. These methods can be viewed as our ``base-cold'' baseline with different inputs and model structures without using ad IDs. They can certainly improve the cold-start performance comparing to classic methods that do not use those features. But they do not consider the ad ID neigher do they optimize to improve the warm-up phase. On the contrary, our method not only uses all available features, but also aims to improving both cold-start and warm-up performance.

For online recommendation, a number of studies aid to improve online learning and accelerate model fitting with incremental data. To name some, \cite{sarwar2002incremental,takacs2008investigation} adjust matrix factorization factors with incremental data; \cite{xu2017r} uses a rating comparison strategy to learn the latent profiles. These methods can learn faster with a small amount of data, but they could not be directly applied if there is no sample for new ads. Moreover, these methods are mostly designed for matrix factorization, so it is not straight-forward to apply them to deep feed-forward models. Different from these methods, Meta-Embedding improve CTR prediction in both the cold-start and warm-up phase, and is designed for deep model with an Embedding \& MLP structure.

Dropout-Net \cite{volkovs2017dropoutnet} handles missing inputs by applying Dropout to deep collaborative filtering models, which can be viewed as a successful training method for pre-training the base models. But since this paper focuses on general CTR predictions than collaborative filtering, we did not include it in our experiments.
\subsubsection*{B. Meta-Learning} It learns how to learn new tasks by using prior experience with related tasks~\cite{vilalta2002perspective,finn2017model,vanschoren2018meta}.  It led to an interest in many areas, such as recommendations~\cite{vartak2017meta}, natural language processing \cite{xu2018lifelong, kiela2018dynamic}, and computer vision \cite{choi2017deep,finn2017model}.

We study how to warm up cold-start ads. In meta-learning literature, this problem relates to few-shot learning~\cite{lake2015human} and fast adaptation~\cite{finn2017model,grant2018recasting}. Specifically, we follow similar spirit of MAML~\cite{finn2017model}, a gradient-based meta-learning method which learns shared model parameters across tasks and achieves fast adaptation towards new tasks. However, MAML cannot be applied directly to CTR prediction because it is designed to learn one model per task, which is unacceptable if there is millions of tasks (ads). 

A recent work of Vartak et al.~\cite{vartak2017meta} also utilizes meta-learning for item cold-start. It represents each user by an averaged representation of items labeled by the user previously. Our approach does not model the user activity and only focus on learning to learn ID embeddings for items (ads). 

In natural language processing, there is another so-called ``Meta-Embedding'' that learns word-embeddings for a new corpus by aggregating pre-trained word vectors from prior domains \cite{yin2016learning,xu2018lifelong,kiela2018dynamic}. It is specific for NLP and do not has common spirit with ours.

\section{Conclusion and Discussion}
In this paper, we proposed a meta-learning approach to address the cold-start problem for CTR predictions. We propose Meta-Embedding, which focuses on learning how to learn the ID embedding for new ads, so as to apply to state-of-the-art deep models for CTR predictions. Built on the pre-trained base model, we suggest a two-phased simulation over previously learned ads to train the Meta-Embedding generator. We seek to improve both cold-start and warm-up performance by leveraging a unified loss function for training. Afterward, when testing, the trained generator can initialize the ID embeddings for new ads, which yields significant improvement in both cold-start and warm-up phases. To the best of our knowledge, it is the first work aiming to improve CTR prediction for both cold-start and warm-up phases for new ads under a general online supervised learning framework.
We verified Meta-Embedding on three real-world datasets over six state-of-the-art CTR prediction models. Experiments showed that, by replacing trivial random initialization with Meta-Embedding, the cold-start and warm-up performances can be significantly improved.

This is an interesting line of work to connect representation learning with learning to learn, especially for online learning tasks including advertising and recommendations. This paper focuses on learning to learn the embedding of ad IDs, which is a specific component of knowledge representation for online CTR/CVR predictions. For the future work, this insight can also be extended to other tasks, for example, learning to learn the model when the distribution of features and labels evolutes over time, or learning to tune the hyper-parameters for specific sub-tasks.
\section{Acknowledgements}
This work is partially supported by the National Key Research and Development Program of China under Grant No. 2017YFB1002104, the National Natural Science Foundation of China under Grant No. U1811461, 61602438, 91846113, 61573335, CCF-Tencent Rhino-Bird Young Faculty Open Research Fund No. RAGR20180111. This work is also funded in part by Ant Financial through the Ant Financial Science Funds for Security Research. Xiang Ao is also supported by Youth Innovation Promotion Association CAS. 

Pingzhong Tang was supported in part by the National Natural Science Foundation of China Grant 61561146398, a China Youth 1000-talent program and an Alibaba Innovative Research program.
\clearpage
\bibliographystyle{ACM-Reference-Format}
\bibliography{ME-bib}

\end{document}